\documentclass[runningheads]{llncs}
\usepackage[T1]{fontenc}
\usepackage{amsmath}
\usepackage{tabularx}

\usepackage{graphicx}
\begin{document}
\title{Cross-domain Fiber Cluster Shape Analysis for Language Performance Cognitive Score Prediction}
\titlerunning{Shape Analysis for Language Performance Prediction}
%
\author{**************************************************\inst{1}}
\author{Yui Lo \inst{1,2,4} \and Yuqian Chen \inst{1,2} \and Dongnan Liu \inst{4} \and Wan Liu\inst{5} \and Leo Zekelman\inst{2,7} \and Fan Zhang\inst{6} \and Yogesh Rathi\inst{1,2} \and Nikos Makris\inst{1,3} \and Alexandra J. Golby\inst{1,2} \and Weidong Cai\inst{4} \and Lauren J. O’Donnell\inst{1,2}}
\authorrunning{Lo et al.}
%
\institute{********************\\
\email{********************@*****}}
\institute{Harvard Medical School, Boston, USA \and
Brigham and Women’s Hospital, Boston, USA\\ \and Massachusetts General Hospital, Boston, USA \and The University of Sydney, Sydney, Australia \and Beijing Institute of Technology, Beijing, China \and University of Electronic Science and Technology of China, Chengdu, China \and Harvard University, Boston, USA\\
\email{odonnell@bwh.harvard.edu}}
\maketitle              
\begin{abstract}
Shape plays an important role in computer graphics, offering informative features to convey an object's morphology and functionality. Shape analysis in brain imaging can help interpret structural and functionality correlations of the human brain. In this work, we investigate the shape of the brain’s 3D white matter connections and its potential predictive relationship to human cognitive function. We reconstruct fiber clusters as sequences of 3D points using diffusion magnetic resonance imaging (dMRI) tractography. To describe each connection, we extract 12 shape descriptors in addition to traditional dMRI connectivity and tissue microstructure features. We introduce a novel framework, \textit{Shape-Fused Fiber cluster Transformer} (SFFormer), that leverages a multi-head cross-attention feature fusion module to predict subject-specific language performance based on dMRI tractography. We assess the performance of the method on a large dataset including 1065 healthy young adults. The results demonstrate that both the transformer-based SFFormer model and its inter/intra feature fusion with shape, microstructure, and connectivity are informative, and together, they improve the prediction of subject-specific language performance scores compared to conventional models. Overall, our results indicate that the shape of the brain’s connections is predictive of human language function.

\end{abstract}

\section{Introduction}
\label{sec:intro}

The study of 3D shape has long been recognized as crucial for computer graphics and medical image analysis \cite{Zhang2004-ob}. In the field of magnetic resonance imaging (MRI), the study of shape has enabled detailed analyses of the folding of the brain’s cortex and the morphology of subcortical gray matter structures \cite{Fischl2012-ss}. However, the shape of the brain’s white matter connections, which transmit information throughout the brain, has been much less studied. 

Diffusion MRI (dMRI) tractography is a unique method that enables the 3D reconstruction of the brain’s white matter connections based on water diffusion in brain tissue \cite{Basser2000-yt}. dMRI tractography produces sequences of 3D points, called streamlines, which can be grouped to define individual brain connections or fiber clusters that have different anatomical shapes (Fig. 1). Quantitative analyses of fiber clusters include tissue microstructure (using water diffusion in tissue), brain connectivity (strength of each connection), and shape analyses. Measures of shape capture white matter variability across individuals \cite{Yeh2020-tb} and changes in aging \cite{Schilling2022-qk}. However, the functional importance of the shape of white matter connections is not well understood. To assess whether fiber cluster shape may be related to language performance, in this work we employ a testbed task of predicting individual language performance. We assess whether the integration of information across shape, microstructure, and connectivity feature domains can enhance the prediction of individual language performance.

\begin{figure}[ht]
  \centering
  \includegraphics[width=0.5\textwidth]{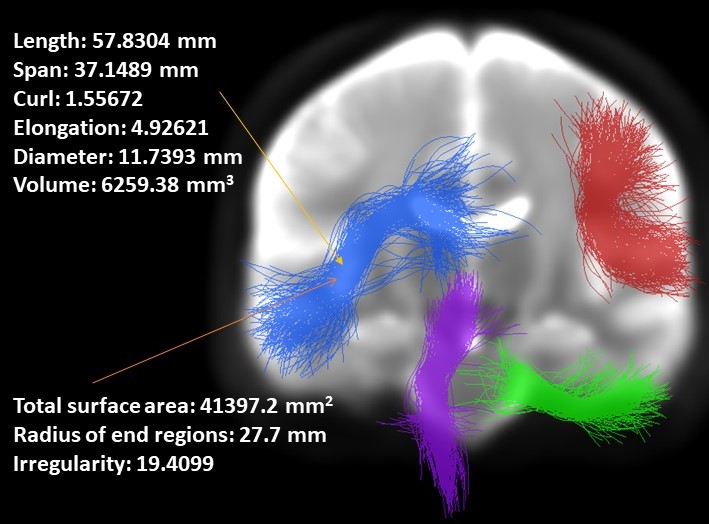}
  \caption{Four example individual white matter connections (fiber clusters) extracted from the entire white matter of the human brain using a fiber clustering approach \cite{Zhang2018-iw}. Example shape descriptors are extracted for the blue fiber cluster.}
\end{figure}

\subsection{Related Work}

In this section, we first give an overview of methods that have been proposed for the prediction of individual language function using dMRI tractography data, then we briefly describe the deep learning techniques upon which our current framework is built.

In the literature, several approaches have been proposed to predict individual cognitive and/or language functional performance using dMRI tractography data \cite{Chen2023-lb,Gong2021-hp,Jeong2021-as,Liu2023-tc}. Tissue microstructure measures derived from dMRI have been shown to relate to language function using traditional (non-deep learning) regression analysis \cite{Zekelman2022-bn}. The studied measures included the fractional anisotropy (FA), which describes the anisotropy of water diffusion within brain tissue, and the mean diffusivity (MD), which describes the overall magnitude of water diffusion \cite{ODonnell2011-bx}, as well as the number of streamlines (NoS), which is thought to relate to the connectivity of the brain \cite{Zhang2022-eb}. In contrast to these traditional features, fiber clusters can be described by shape measures such as surface area and volume, as well as recently proposed, fiber-tract-specific measures such as the surface area of the region where the tract inserts in to the gray matter \cite{Yeh2020-tb}. Other shape measures have been proposed for tractography, including curvature and torsion \cite{Batchelor2006-rt}, fiber dispersion \cite{Savadjiev2010-yj}, and volume \cite{Lebel2012-xz}.

Recent deep-learning methods have investigated the prediction of individual language performance using dMRI tractography. A convolutional neural network (CNN) based deep learning method has shown that connectivity is predictive of language proficiency in children with epilepsy \cite{Jeong2021-as}. A geometric deep-learning approach showed that microstructure and connectivity are predictive of language function in healthy young adults \cite{Chen2023-lb}. A multilayer network approach was applied to predict language performance in human aging \cite{Feng2022-he}. In contrast with these methods, we focus on a novel transformer-based network design.

In recent years, transformer models \cite{Gorishniy2021-yw} are increasingly popular for computer vision tasks such as object detection \cite{Zhu2020-sk}, classification \cite{Dosovitskiy2020-hw}, and segmentation \cite{Strudel2021-qw}. The advantage of transformers over CNNs is the use of multi-head self-attention \cite{Vaswani2017-nm} to enhance the model's ability to interpret complex semantic and structural feature relationships more comprehensively. Transformers have also been shown to be successful in many medical image applications, including dMRI \cite{Tiwari2023-bv,Chen2022-bj,Yang2023-qc,Zhang2022-vi}. There is a substantial body of literature on transformer models to predict tissue microstructure, including SwinDTI \cite{Tiwari2023-bv}, Microstructure Estimation Transformer with Sparse Coding \cite{Zheng2023-xv}, Hybrid Graph Transformer (HGT) \cite{Chen2022-bj}, and 3D HGT \cite{Yang2023-qc}. Applications of transformers in tractography analysis are relatively limited, such as TractoFormer \cite{Zhang2022-vi} for whole-brain tractography analysis. Consequently, it is of interest to investigate the application of transformers in the analysis of tractography data.

\section{Methodology}

\begin{figure*}[ht]
  \centering
  \includegraphics[width=\textwidth]{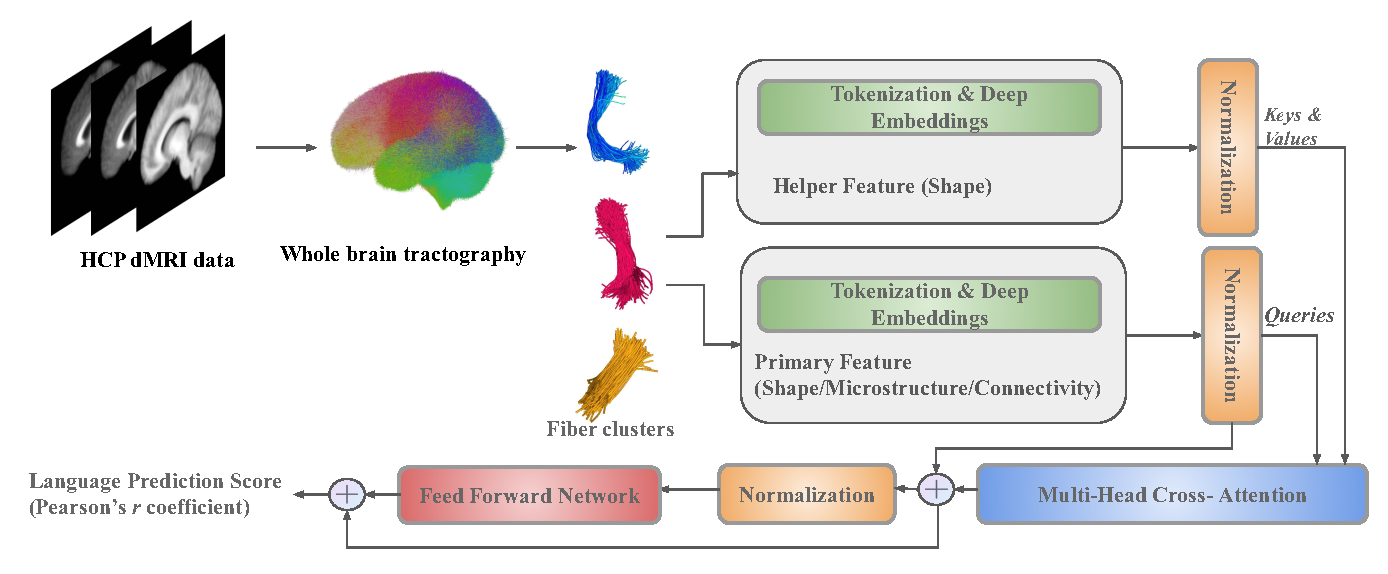}
  \caption{Overview of the SFFormer framework. HCP dMRI data undergoes whole brain tractography to obtain 953 fiber clusters. The microstructure, connectivity and shape features of the fiber clusters are calculated and used as inputs to the SFFormer framework that leverages both a helper feature and a primary feature in the multi-head cross-attention module to output a language prediction score.}
\end{figure*}
\subsection{Tractography and Fiber Clustering}
In this work, we study the shape of the fiber clusters of 1065 healthy young adults (575 females and 490 males, 28.7 years old on average) from the Human Connectome Project Young Adult (HCP-YA) dataset \cite{Van_Essen2013-bz,Van_Essen2012-qh}. Whole brain tractography is generated for each subject’s dMRI data using a two-tensor unscented Kalman filter method \cite{Malcolm2010-zl} that can represent multiple crossing fibers, enabling anatomically sensitive estimation of the pathway and connectivity of fiber clusters \cite{He2023-yo}. Tractography is then parcellated into 953 fiber clusters using an anatomically curated tractography brain atlas \cite{Zhang2018-iw}. Each fiber cluster contains hundreds of streamlines and represents a particular connection in the human brain (Fig. 1). 

\subsection{Traditional and Shape Features}
For each cluster, we compute traditional tissue microstructure features including fractional anisotropy (FA) and mean diffusivity (MD), and the traditional connectivity feature of the number of streamlines (NoS) \cite{Zhang2022-eb}. These features are used to compare and evaluate the shape features. 

We study 12 fiber cluster shape features that are considered to provide a comprehensive shape analysis of tractography \cite{Yeh2020-tb}. Features include length, diameter, elongation, span, curl, volume, trunk volume, branch volume, total surface area, total radius of end regions, total area of end regions, and irregularity. These shape features are computed for all fiber clusters from all subjects by applying the software DSIStudio \cite{Yeh2020-tb}. Full definitions and calculations of the shape measures are presented in DSIStudio \cite{Yeh2020-tb}. 

\subsection{Shape-Fused Fiber Cluster Transformer (SFFormer)}

In this section, we present our proposed SFFormer for subject-specific language score prediction. As depicted in Fig. 2, the SFFormer model comprises a tokenization module and an encoder-only transformer architecture, specifically tailored for prediction tasks. This encoder-only design aligns with the task of focusing on learning the fiber clusters’ features for language score predictive outcomes. The SFFormer encoder comprises a stack of 1-4 identical layers. Each layer includes a multi-head attention module and a feed-forward network. 

The tokenization module \cite{Gorishniy2021-yw} performs deep embedding of a particular feature (e.g., FA or length) of dimension 1 $\times$ 953. To create the embedding, we multiply the input data ($x$) with random initialized weights and then add random initialized biases. This process prepares the data for the multi-head cross-attention module in the deep learning pipeline.

We extended our design from the vanilla transformer \cite{Gorishniy2021-yw}. We naturally take a fiber cluster feature as a token to utilize the long-range dependency of all cluster features to benefit prediction. We employ a multi-head mechanism \cite{Vaswani2017-nm} that is well suited for processing long sequences, such as the 953 fiber cluster features, because each head independently attends to different parts of the input sequence.

We design a multi-head cross-attention module to fuse features from shape, microstructure, and connectivity feature domains. Instead of using the transformer’s self-attention mechanism, our multi-head cross-attention module can fuse the features of different domains to symmetrically combine two embedding sequences of the same dimension, where one sequence is used as the query (Q) input. The other sequence is used as the key (K) and value (V) inputs in SFFormer to provide feature fusion. As it requires two embeddings SFFormer captures and attends to information from different features simultaneously. The motivation is to more effectively determine varying attention weights by utilizing the dual-stream input framework. This methodology emphasizes cross-attention to concurrently train on the primary feature to attempt to integrate key information from both data streams.

\subsection{Implementation Details}
Our model is trained and tuned with Optuna Hyperparameters \cite{Akiba2019-gm}, set to 20 trials. The model is configured with the ReGLU activation and the He initialization \cite{He2015-zi} with 8 attention heads. The model is trained and evaluated with batch sizes of 8 for 1000 epochs with patience of 50 epochs. All of the experiments are split into three-fold cross-validation. The training is optimized with Adam \cite{Kingma2014-lm}, where the learning rate is set between 1e-5 and 1e-3 with a log uniform weight decay between 1e-6 and 1e-3. The tokens are set between 64 to 512 with larger embeddings capturing more information. The dropouts for attention and feed-forward modules are set between 0 and 0.5, and 0 and 0.2 for residual connections. All experiments are conducted on an NVIDIA RTX A5000 GPU using PyTorch 1.7.1 \cite{Paszke2019-nb}.

\section{EXPERIMENTS \& RESULTS}
First, we conduct experiments to perform subject-specific language score prediction based on individual features. We compare the performance of a state-of-the-art 1DCNNN model \cite{Liu2023-tc} and a baseline transformer model, when trained on an individual microstructure, connectivity, or shape feature. Next, we fuse each feature with a selected helper shape feature and apply the SFFormer model. The helper shape feature is selected as the best-performing shape feature when using the baseline transformer model. We select diameter as the helper shape feature for TPVT score prediction and irregularity as the helper shape feature for TORRT score prediction. 

\subsection{Language Assessments Scores}
We predict subject-specific performance on two language assessments provided by HCP-YA, including the NIH Toolbox Picture Vocabulary Test (TPVT) and the NIH Toolbox Oral Reading Recognition Test (TORRT) \cite{Gershon2014-nu,Weintraub2013-es}. TPVT measures vocabulary comprehension and is a receptive language assessment \cite{Gershon2014-nu}. TORRT measures reading decoding and is a spoken language assessment \cite{Gershon2014-nu}.

\subsection{Evaluation Metric}
The Pearson correlation coefficient (\textit{r}) \cite{Sedgwick2012-dg} is employed to evaluate language performance prediction as it is a prevalent metric in neurocognitive performance prediction \cite{Gong2021-hp,Kim2021-za,Tian2021-ci}. Pearson’s r measures the strength and direction (positive or negative) of the linear association between two variables.

\begin{table}[t]
\begin{center}
\caption{Prediction performance for TPVT (\textit{r}). Shape features shown in italics outperform the best-performing traditional feature. Bolded features show the best performance across the three different learning models.} 
\begin{tabularx}{\textwidth}{|X|X|X|X|}
  \hline
  Features & CNN \cite{Liu2023-tc} & Vanilla Transformer & SFFormer (helper: diameter)
  \\
  \hline
  \multicolumn{4}{|X|}{Microstructure} \\
  \hline
  FA & 0.293±0.063 & \textbf{0.418±0.077} & 0.404±0.079 \\ \hline
  MD & 0.260±0.041 & 0.337±0.098 & \textbf{0.338±0.098} \\ \hline
  \multicolumn{4}{|X|}{Connectivity} \\ \hline
  NoS & 0.395±0.054 & 0.410±0.103 & \textbf{0.417±0.007} \\ \hline
  \multicolumn{4}{|X|}{Shape} \\
  \hline
Length  &  0.133±0.039 & 0.330±0.079 & \textbf{0.414±0.080} \\ \hline
Span & 0.119±0.044 & 0.355±0.094 & \textbf{0.417±0.098} \\ \hline
Curl & 0.203±0.092 & 0.310±0.070 & \textbf{0.407±0.081} \\ \hline
Volume & 0.381±0.063 & 0.410±0.102 & \textbf{\textit{0.423±0.071}} \\ \hline
Trunk Volume & 0.156±0.083 & 0.275±0.041 & \textbf{0.414±0.084} \\ \hline
Branch Volume & 0.376±0.064 & 0.414±0.096 & \textbf{\textit{0.430±0.079}} \\ \hline
Diameter & 0.406±0.082 & \textbf{0.419±0.083} &  \textbf{-----------------------}\\ \hline
Elongation & 0.313±0.070 & 0.392±0.074 & \textbf{\textit{0.419±0.083}} \\ \hline
Total surface area & 0.395±0.060 & \textbf{0.418±0.098} & 0.406±0.092 \\ \hline
Radius of end regions & 0.235±0.045 & 0.347±0.125 & \textbf{\textit{0.429±0.084}} \\ \hline
Surface area of end regions & 0.406±0.080 & 0.414±0.100 & \textbf{\textit{0.418±0.092}} \\ \hline
Irregularity & 0.322±0.041 & 0.391±0.092 & \textbf{0.416±0.071} \\ \hline
\end{tabularx}
\end{center}
\end{table}
\subsection{Results and Discussions}
Tables 1 and 2 show the performance of the three compared models for predicting subject-specific vocabulary comprehension (TPVT) and subject-specific oral reading (TORRT) scores, respectively. 

\begin{table}[h]
\begin{center}
\caption{Prediction performance for TORRT (\textit{r}). Shape features shown in italics outperform the best-performing traditional feature. Bolded features show the best performance across the three different learning models.} 
\begin{tabularx}{\textwidth}{|X|X|X|X|}
  \hline
  Features & CNN \cite{Liu2023-tc} & Vanilla Transformer & SFFormer (helper: irregularity)
  \\
  \hline
  \multicolumn{4}{|X|}{Microstructure} \\
  \hline
  FA & 0.332±0.055 & 0.382±0.059 & \textbf{0.383±0.06} \\ \hline
  MD  & 0.315±0.004  & 0.344±0.021  & \textbf{0.374±0.06} \\ \hline
  \multicolumn{4}{|X|}{Connectivity} \\ \hline
  NoS & 0.349±0.024 & 0.345±0.061 & \textbf{0.372±0.05} \\ \hline
  \multicolumn{4}{|X|}{Shape} \\
  \hline
Length &  0.103±0.002 &  0.301±0.056 &  \textbf{0.376±0.053}  \\ \hline
Span &  0.126±0.017 &  0.318±0.071 &  \textbf{0.377±0.072}  \\ \hline
Curl &  0.241±0.014 &  0.285±0.061 &  \textbf{0.377±0.075} \\ \hline
Volume &  0.324±0.016 &  \textbf{\textit{0.392±0.083}} &  0.379±0.066  \\ \hline
Trunk Volume &  0.184±0.035 &  0.260±0.039 &  \textbf{\textit{0.384±0.123}}  \\ \hline
Branch Volume &  0.357±0.021 &  \textbf{0.377±0.075} &  0.362±0.073  \\ \hline
Diameter &  0.315±0.038 &  0.390±0.071 &  \textbf{\textit{0.398±0.050}}  \\ \hline
Elongation &  0.275±0.005 &  0.363±0.045 &  \textbf{\textit{0.376±0.049}}  \\ \hline
Total surface area &  0.368±0.046 &  \textbf{0.391±0.079} &  0.369±0.056  \\ \hline
Radius of end regions&  0.3196±0.063 &  0.341±0.087 &  \textbf{0.374±0.053}  \\ \hline
Surface area of end regions &  0.330±0.001 &  \textbf{0.374±0.085} &  0.371±0.062  \\ \hline
Irregularity &  0.341±0.021 &  \textbf{\textit{0.439±0.062}} &  \textbf{-----------------------}  \\ \hline
\end{tabularx}
\end{center}
\end{table}

The CNN model \cite{Liu2023-tc}, shown in the second column of Tables 1 and 2, successfully predicts language performance, though it is outperformed by both the baseline transformer and SFFormer models. When using the CNN model, the NoS feature is the most informative traditional feature, while several shape features (shown in italics) outperform NoS.

The baseline transformer model (third column of Tables 1 and 2) outperforms the CNN model for all input features. This indicates that the transformer improves the performance of the language score prediction task. The FA feature is the most informative traditional feature. Multiple shape features (shown in italics) outperform FA, including diameter (Table 1), volume, diameter, total surface area, and irregularity (Table 2).

The SFFormer (fourth column of Tables 1 and 2) successfully predicts language performance, and most of its features outperform the baseline model as well as the state-of-the-art CNN model. This indicates that the domain fusion technique effectively contributes to subject-specific language score prediction, where the performance improves when information from the helper feature is included to aid the overall training. In Table 1, shape features have comparable or better performance than FA and MD, with various shape features (shown in italics), including the surface area of end regions, elongation, volume, radius of end regions, and branch volume, surpassing the traditionally best-performing feature, NoS, in predicting language performance. Also, Table 2 reveals that FA is the most informative traditional feature, and FA outperforms several shape features, such as trunk volume and diameter (shown in italics).

In summary, the evaluation presented in Tables 1 and 2 demonstrates the superior predictive power of shape features and domain fusion in the SFFormer model, marking an improvement over traditional features and surpassing the state-of-the-art methods of the CNN model and the self-attention vanilla transformer model. Future work may explore models that incorporate multiple input features.

\section{Conclusion}
In this paper, we proposed the SFFormer, which utilizes a multi-head cross-attention module to fuse features from different domains to improve the prediction results. Our SFFormer results show that measures of the shape of fiber cluster connections are informative for the prediction of individual, subject-specific language performance. The evaluation of the HCP-YA dataset suggests inter/intra domain feature fusion to be beneficial towards better prediction. This suggests that shape-related features are useful for predicting and evaluating various cognitive abilities, potentially outperforming microstructural and connectivity features in certain scenarios. Overall, this suggests that the shape of the white matter fiber clusters relates to important functions of the human brain.

\section{COMPLIANCE WITH ETHICAL STANDARDS}
This study uses public HCP imaging data; no ethical approval was required.

\section{ACKNOWLEDGEMENTS}
This work is in part supported by the National Key R\&D Program of China (No. 2023YFE0118600), and the National Natural Science Foundation of China (No. 62371107). This work is supported by the University of Sydney International Scholarship
%
%
\bibliographystyle{splncs04}
\bibliography{reference}
%




\end{document}